\documentclass[11pt,a4paper]{article}
\usepackage[hyperref]{acl2018}
\usepackage{times}
\usepackage{latexsym}
\usepackage{booktabs}
\usepackage{url}
\usepackage{graphicx}
\usepackage{amsthm}
\usepackage{multirow}

\aclfinalcopy 

\title{Generating Descriptions for Sequential Images  with Local-Object Attention and Global Semantic Context Modelling}

\author{Jing Su\textsuperscript{1}, Chenghua Lin\textsuperscript{2}, Mian Zhou\textsuperscript{3}, Qingyun Dai\textsuperscript{4}, Haoyu Lv\textsuperscript{4} \\
\textsuperscript{1}Guangdong Ocean University, \textsuperscript{2}University of Aberdeen \\
\textsuperscript{3} Tianjin University of Technology, \textsuperscript{4} Guangdong University of Technology \\
  {\tt jingsuw@163.com,chenghua.lin@abdn.ac.uk} \\
  {\tt zhoumian@tjut.edu.cn,1144295091@qq.com,lvhaoyuchn@163.com} 
}

\date{}

\begin{document}
\maketitle
\begin{abstract}
In this paper, we propose an end-to-end CNN-LSTM model for generating descriptions for sequential images with a local-object attention mechanism. To generate coherent descriptions, we capture global semantic context using a multi-layer perceptron, which learns the dependencies between sequential images. A paralleled LSTM network is exploited for decoding the sequence descriptions. 
Experimental results show that our model outperforms the baseline across three different evaluation metrics on the datasets published by Microsoft.
\end{abstract}

\section{Introduction}

Recently, automatically generating image descriptions has attracted considerable interest in the fields of computer vision and nature language processing. Such a task is easy to humans but highly non-trivial for machines as it requires not only capturing the semantic information from images (e.g., objects and actions) but also needs to generate human-like natural language descriptions.

Existing approaches to generating image description are dominated by neural network-based methods, which mostly focus on generating description for a single image \cite{Karpathy2015Deep,Xu2015Show,Jia2015Guiding,You2016Image}. Generating descriptions for sequential images, in contrast, is much more challenging, i.e., the information of both  individual images as well as the dependencies between images in a sequence needs to be captured. 

\citet{Huang2016} introduce the first sequential vision-to-language dataset and exploit  Gated Recurrent Units (GRUs) \cite{Cho2014Learning} based encoder and decoder for the task of visual storytelling. However, their approach only considers image information of a sequence at the first time step of the decoder, where the local attention mechanism is ignored which is important for capturing the correlation between the features of an individual image and the corresponding words in a description sentence. \citet{Yulicheng2017} propose a hierarchically-attentive Recurrent Neural Nets (RNNs) for album summarisation and storytelling. To generate descriptions for an image album, their hierarchical framework selects representative images from several image sequences of the  album, where the selected images might not necessary have correlation to each other. 

\begin{figure*}
\centering 
\includegraphics[width = 0.8\textwidth]{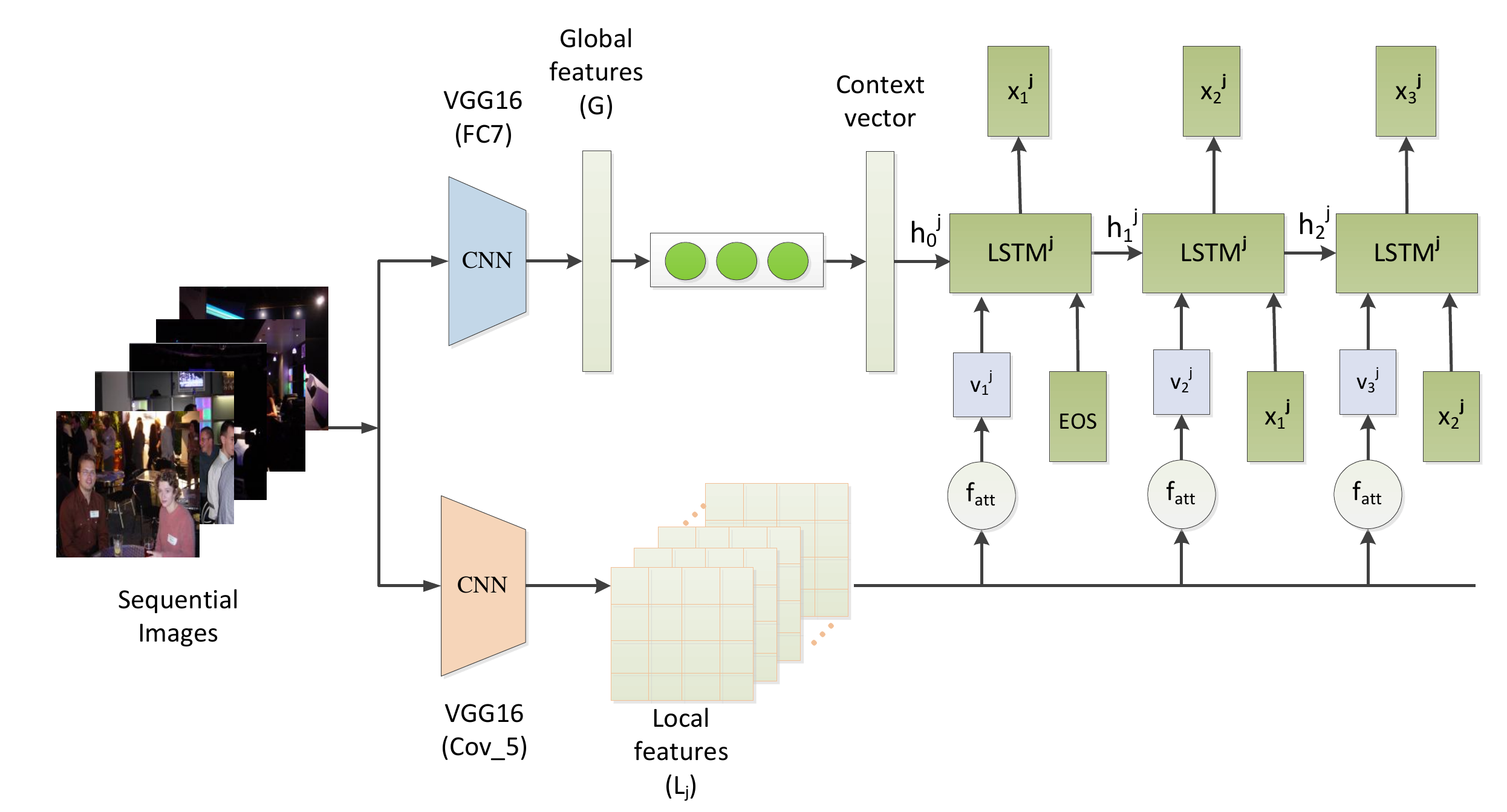}
\caption{The architecture of our CNN-LSTM model with global semantic context.} \label{fig:fram}
\end{figure*}

In this paper, we propose an end-to-end CNN-LSTM model with a local-object attention mechanism for generating story-like descriptions for multiple images of a sequence. To improve the coherence of the generated descriptions, we exploit a  paralleled long short-terms memory (LSTM) network and learns global semantic context by embedding the global features of sequential images as an initial input to the hidden layer of the LSTM model. 
We evaluate the performance of our model on the task of generating story-like descriptions for an image sequence on the sequence-in-sequence (SIS) dataset published by Microsoft. We hypothesise that by taking into account global context, our model can also generate better descriptions for individual images. Therefore, in another set of experiments, we further test our model on the  Descriptions of Images-in-Isolation (DII) dataset for generating descriptions for each individual image of a sequence. 
Experimental results show that our model outperforms a baseline developed based on the state-of-the-art image captioning model~\cite{Xu2015Show} in terms of BLEU, METEOR and ROUGE, and can generate sequential descriptions which preserve the dependencies between sentences.  

\section{Related Work}

Recent successes in machine translation using Recurrent Neural Network (RNN) \cite{Bahdanau2014Neural,Cho2014Learning} catalyse the adoption of neural networks in the task of image caption generation. Early works of image caption generation based on CNN-RNN networks have been made great progress.\citet{Vinyals2014Show} propose an encoder-decoder model which utilises a Convolutional Neural Network (CNN) for encoding the input image into a vector representation and a Recurrent Neural Network (RNN) for decoding the corresponding text description. Similarly, \citet{Karpathy2015Deep} present an alignment model based on a CNN and a bidirectional RNN which can align segment regions of an image to the corresponding words of a text description. 
\citet{Donahue2014} propose a Long-term Recurrent Convolutional Network (LRCN) which integrates convolutional layers and long-range temporal recursion for generating image descriptions.

Recently, the attention mechanism \cite{Xu2015Show,You2016Image,Lu16,Zhou2016} has been widely used and proved to be effective in the task of image description generation. For instance, \citet{Xu2015Show} explore two kinds of attention mechanism for generating image descriptions, i.e., soft-attention and hard-attention, whereas \citet{You2016Image} exploits a selective semantic attention mechanism for the same task. 

There is also a surge of research interest in visual storytelling \cite{Kim2014,Sigurdsson2016,Huang2016,Yulicheng2017}. \citet{Huang2016} collect stories using Mechanical Turk and translate a sequence of images into story-like descriptions by extending a GRU-GRU framework. \citet{Yulicheng2017} utilise a hierarchically-attentive structures with combined RNNs for photo selection and story generation. However, the above mentioned approaches for generating descriptions of sequential images do not explicitly capture the dependencies between each individual images of a sequence, which is the gap that we try to address in this paper.

\section{Methodology}

In this section, we describe the proposed CNN-LSTM model with local-object attention. In order to generate coherent descriptions for an image sequence, we introduce global semantic context and a paralleled LSTM in our framework as  shown in Figure.~\ref{fig:fram}. Our model works by first extracting  the global features of sequential images using a CNN network (VGG16) \cite{Simonyan2014Very}, which has been extensively used in image recognition. Here a VGG16 model contains 13 convolutional layers, 5 pooling layers and 3 fully connected layers. The extracted global features are then embedded into a global semantic vector with a multi-layer perceptron as the initial input to the hidden layer of a paralleled LSTM model. Our model then applies the last convolutional-layer operation from the VGG16 model to generate the local features of each image in sequence. Finally, we introduce a paralleled  LSTM model and a local-object attention mechanism to decode sentence descriptions.

\subsection{Features Extraction and Embedding}
Sequential image descriptions are different from single image description due to the spatial correlation between images. Therefore, in the encoder, we exploit both global and local features for describing the content of sequential images. We extract global features of the sequential images with the second fully connected layer (FC7) from VGG16 model. The global features are denoted by $G$ which are a set of 4096-dimension vectors. Then, we select the features of the final convolutional layer ($Cov\_5$) from the VGG16 model to represent local features for each image in the sequence. The local features are denoted as $L_j$ ($j=1$,$\dots$,$N$), where $N$ is the number of images in the sequence. In our experiment, we follow \citet{Huang2016} and set 5 as the number of images in a sequence. Finally, we embed the global features $G$ into a 512-dimension context vector via a multi-layer perceptron which is then used as the initial input of the hidden layer in LSTM model.

\subsection{Sequential Descriptions Generation}

In the decoding stage, our goal is to obtain the most likely text descriptions of a given sequence of images. This can be generated by training a model to maximize the log likelihood of a sequence of sentences $S$, given the corresponding sequential images $I$ and the model parameters $\theta$, as shown in Eq.~\ref{eq:probability}. 

\begin{equation} \label{eq:probability}
\theta^* = \mathop{\arg\max}_{\theta} \sum_{j=1}^N \sum_{(I,s_j)} log\, p(s_j|I,\theta)
\end{equation} 
Here $s_j$ denotes a sentence in $S$, and $N$ is the total number of sentences in $S$.

Assuming a generative model of each sentence $s_j$  produces each word in the sentence in order, the log probability of $s_j$ is given by the sum of the log probabilities over the words:
\begin{equation}
\log p(s_j|I) = \sum_{t=1}^C log\, p(s_{j,t}|I,s_{j,1},s_{j,2}...s_{j,t-1})
\end{equation}
where $s_{j,t}$ represents the $t^{th}$ word in the $j^{th}$ sentence and $C$ is the total number of words of $s_j$. 

We utilize a LSTM network \cite{Hochreiter1997Long} to produce a sequence descriptions conditioned on the local feature vectors, the previous generated words, as well as the hidden state with a global semantic context. Formally, our LSTM model is formulated as follows: 
\begin{eqnarray}
&& i^j_t = \sigma(W_{xi}x^j_{t-1}+W_{hi}h^j_{t-1}+W_{vi}v^j_t+b_i) \nonumber\\
&& f^j_t = \sigma(W_{xf}x^j_{t-1}+W_{hf}h^j_{t-1}+W_{vf}v^j_t+b_f) \nonumber\\
&& o^j_t = \sigma(W_{xo}x^j_{t-1}+W_{ho}h^j_{t-1}+W_{vo}v^j_t+b_o) \nonumber\\
&& q^j_t = \varphi(W_{xq}x^j_{t-1}+W_{hq}h^j_{t-1}+W_{vq}v^j_t+b_q) \nonumber\\
&& c^j_t = f^j_t \odot c^j_{t-1} + i^j_t \odot q^j_t \nonumber\\
&& h^j_t = o^j_t\odot \varphi(c^j_t) \label{eq3}
\end{eqnarray}
where $i^j_t$, $f^j_t$, $o^j_t$ and $c^j_t$ represents input gates, forget gates, output gates and memory, respectively. $q^j_t$ represents the updating information in the memory $c^j_t$. $\sigma$ denotes the sigmoid activation function, $\odot$ represents the element-wise multiplication, and $\varphi$ indicates the hyperbolic tangent function. $W_\bullet$ and $b_\bullet$ are the parameters to be estimated during training. Also $h^j_t$ is the hidden state at time step $t$ which will be used as an input to the LSTM unit at the next time step.

Here, we utilize a multilayer perceptron to model the global semantic context which can be viewed as the initial input of the hidden state $h^j_0$, where every initial value $h^j_0$ in the LSTM model is equal and is defined as:
\begin{equation}
 h^j_0 = W_0 \, \varphi(W_gG+b_g)  
\end{equation}

\begin{figure*}
\centering {\includegraphics[width = 1.0\textwidth]{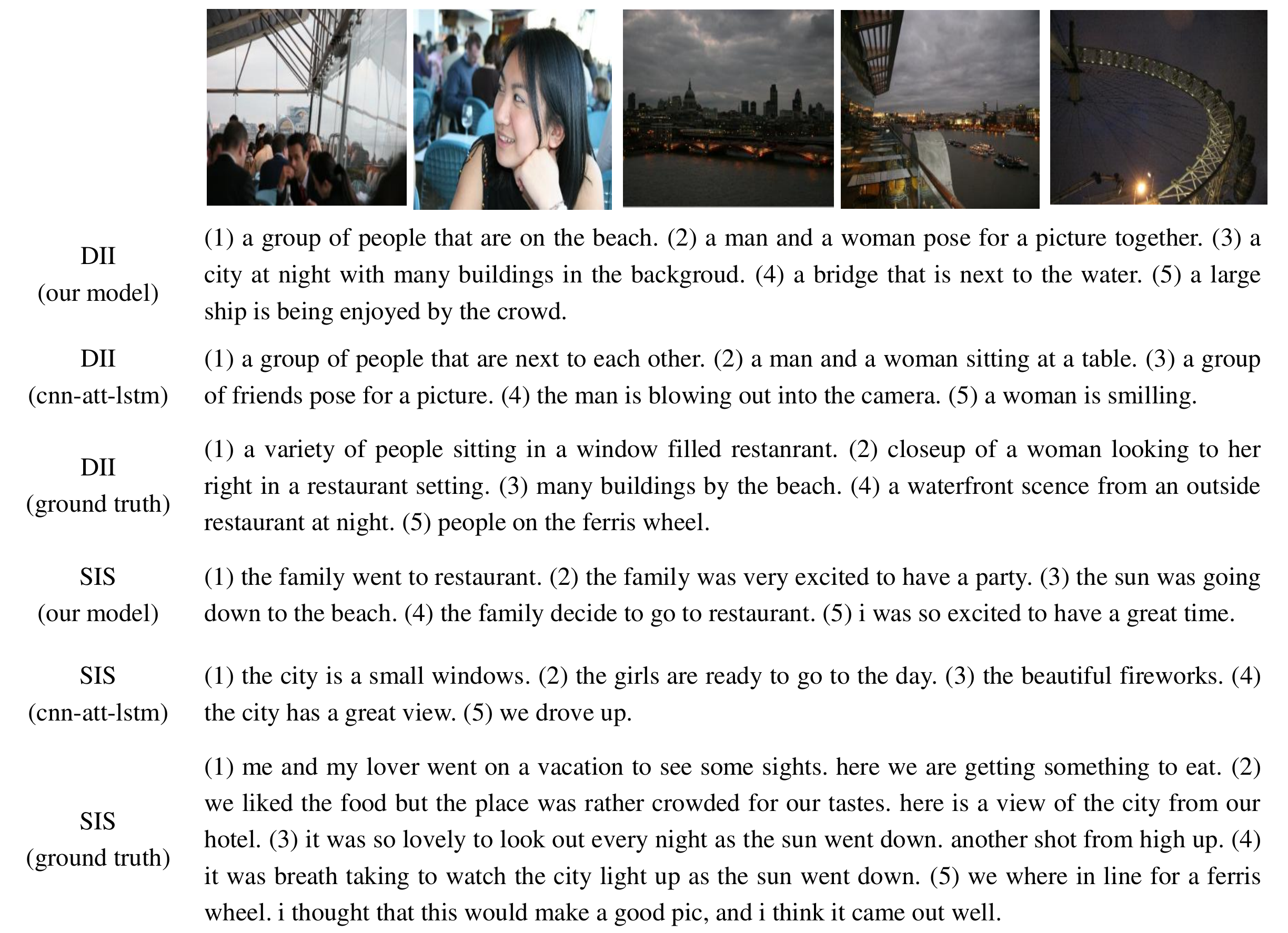}}
\caption{Example of sequential descriptions generated by our model, the baseline, and the ground truth.} \label{fig:finalcaption}
\end{figure*}

When modelling local context, the local context vector $v^j_t$ is a dynamic representation of the relevant part of the $j^{th}$ image in a sequence at time $t$. In Eq.~\ref{eq:attention}, we use the attention mechanism $ f_{att} $ proposed by \cite{Bahdanau2014Neural} to compute the local attention vector $v^j_t$, where the corresponding weight $k^j_t$ of each local features $L_j$ is computed by a softmax function with input from a multilayer perceptron which considers both the current local vector $L_j$ and the hidden state $h^j_{t-1}$ at time $t-1$. 
\begin{equation}
 k^j_t = softmax(W_k\,tanh(W_{lv}L^j+W_{hv}h^j_{t-1}+b_v)) 
\end{equation}
\begin{equation} \label{eq:attention}
 v^j_t = \sum_{i=1}^M k^j_{it} L^j_i 
\end{equation}

\begin{figure*}
\centering {\includegraphics[width = 1.0\textwidth]{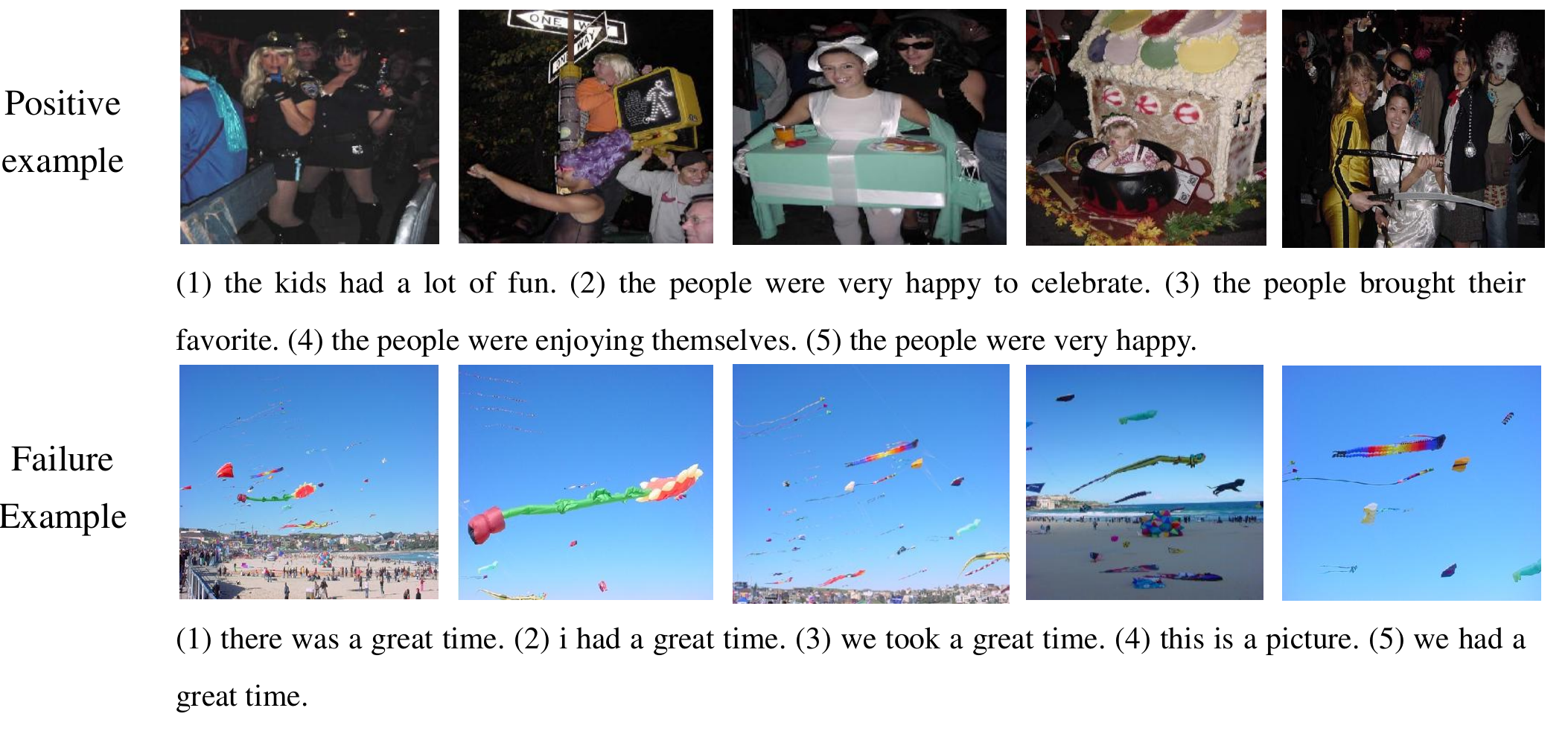}}
\caption{Error analysis of our model. First row: our model generates correct captions. Second row: failure cases due to severe overfitting.} \label{fig:sis_caption}
\end{figure*}

\section{Experiments}

\noindent {\bf Dataset.}~~~

Both the SIS and DII datasets are published by Microsoft\footnote{\url{http://visionandlanguage.net/VIST/}}, which have a similar data structure, i.e., each image sequence consists of five images and their corresponding descriptions. The key difference is that descriptions of SIS consider the dependencies between images, whereas the descriptions of DII are generated for each individual image, i.e., no dependencies are considered. As the full DII and SIS datasets are quite large, we only used part of both datasets for our initial experiments, where the dataset statistics are shown in Table~\ref{tb:dataset}. 

\begin{table}[t!]
\begin{center} \small
\begin{tabular}{|c|c|c|c|} 
    \hline \bf Dataset & \bf Train & \bf Test   & \bf Vocab. Size  \\  
    \hline DII  & 23,415 & 1,665 &  10,000  \\ 
	 \hline SIS  & 110,905  & 10,370  & 18,000    \\
     \hline
\end{tabular}
\end{center}
\caption{\label{tb:dataset} Dataset statistics.}
\end{table}

\noindent {\bf Evaluation.}~~~We compare our model with the sequence-to-sequence baseline (cnn-att-lstm) with attention mechanism \cite{Xu2015Show}.  The cnn-att-lstm baseline only utilises the local attention mechanism which combines visual concepts of an image with the corresponding words in a sentence. Our model, apart from adopting a local-object attention, can further model global semantic context for capturing the correlation between sequential images. 
\begin{table}[t!]
\begin{center} \small
\begin{tabular}{|c|c|ccc|}
    \hline \bf Dataset & \bf Method & \bf BLEU & \bf METEOR & \bf ROUGE\\ 
    \hline 
    \multirow{2}{*}{DII} & cnn-att-lstm &  36.1  & 9.2  & 26.9    \\ 
	     & Our model   & \bf 40.1 & \bf 11.2  & \bf 29.1 \\ 
     \hline  \multirow{2}{*}{SIS}& cnn-att-lstm &  15.2 & 4.6  & 13.6  \\
       & Our model  & \bf 17.2  &  \bf 5.5 & \bf 15.2  \\ 
	\hline
\end{tabular}
\end{center}
\caption{\label{tb:metrics}Evaluation of the quality of descriptions generated for sequential images.}
\end{table}

Table~\ref{tb:metrics} shows the experimental results of our model on the task of generating descriptions for sequential images  with three popular evaluation metrics, i.e. BLEU, Meteor and ROUGE. It can be observed from Table~\ref{tb:metrics} that our model outperforms the baseline on both SIS and DII datasets for all evaluation metrics. It is also observed that the scores of the evaluation metric are generally higher for the DII dataset than the SIS dataset. The main reason is that the SIS dataset contains more sentences descriptions in a sequence and more abstract content descriptions such as ``breathtaking'' and ``excited'' which are difficult to understand and prone to overfitting.

Figure~\ref{fig:finalcaption} shows an example sequence of five images as well as their corresponding descriptions generated by our model, the baseline (cnn-att-lstm), and the ground truth. 
For the SIS dataset, it can observed that our model can capture more coherent story-like descriptions. For instance, our model can learn the social word ``family'' to connect the whole story and learn the emotional words ``great time'' to summarise the description. However, the baseline model failed to capture such important information. 
Our model can learn dependencies of visual scenes between images even on the DII dataset. For example, compared to the descriptions generated by cnn-att-lstm, our model can learn the visual word ``beach'' in image 1 by reasoning from the visual word ``water'' in image 4. 

Our model can generally achieve good results by capturing the global semantics of an image sequence such as the example in the first row of Figure~\ref{fig:sis_caption}. However, our model also has difficulties in generating meaningful descriptions in a number of cases. For instance, our model generates fairly abstractive descriptions such as ``a great time'' due to severe overfitting, as shown in the second row of Figure~\ref{fig:sis_caption}. We suppose the issue of overfitting is likely to be alleviated by adding more training data or using more effective algorithm for image feature extraction.

\section{Conclusion}

In this paper, we present a local-object attention model with global semantic context for sequential image descriptions. Unlike other CNN-LSTM models that only employ a single image as input for image caption, our proposed method can generate descriptions of sequential images by exploiting the global semantic context to learn the dependencies between sequential images. Extensive experiments on two image datasets (DII and SIS) show promising results of our model.

\section*{Acknowledgement}
We thank Ehud Reiter and  Kees van Deemter for their helpful discussion. This paper was also supported partly by Science and Technology Project of Guangdong Province (No.503314759024).

\bibliography{acl2018}
\bibliographystyle{acl_natbib}

\end{document}